\RequirePackage{snapshot}
\documentclass[10pt,twocolumn,letterpaper]{article}
\usepackage{iccv}
\usepackage{times}
\usepackage{epsfig}
\usepackage{graphicx}
\usepackage{amsmath}
\usepackage{amssymb}
\usepackage{booktabs}
\usepackage{multirow}
\usepackage{float}

\graphicspath{{figs/}}

\usepackage[pagebackref=true,breaklinks=true,letterpaper=true,colorlinks,bookmarks=false]{hyperref}

\iccvfinalcopy 

\def\iccvPaperID{9782} 

\ificcvfinal\fi

\makeatletter
\def\thanks#1{\protected@xdef\@thanks{\@thanks
        \protect\footnotetext{#1}}}
\makeatother

\begin{document}

\title{Dynamic Token Pruning in Plain Vision Transformers\\for Semantic Segmentation}

\author{
Quan Tang\textsuperscript{\rm 1*}\thanks{* Equal contribution. \quad\textdagger~Corresponding author: fgliu@scut.edu.cn, yifan.liu04@adelaide.edu.au.}\quad\quad
Bowen Zhang\textsuperscript{\rm 2*}\quad\quad
Jiajun Liu\textsuperscript{\rm 3}\quad\quad
Fagui Liu\textsuperscript{\rm 1\textdagger}\quad\quad
Yifan Liu\textsuperscript{\rm 2\textdagger}
\and
{\textsuperscript{\rm 1}South China University of Technology}\quad\quad
{\textsuperscript{\rm 2}The University of Adelaide}\quad\quad
{\textsuperscript{\rm 3}CSIRO}
}
\maketitle
\ificcvfinal\thispagestyle{empty}\fi

\begin{abstract}
Vision transformers have achieved leading performance on various visual tasks yet still suffer from high computational complexity. The situation deteriorates in dense prediction tasks like semantic segmentation, as high-resolution inputs and outputs usually imply more tokens involved in computations. Directly removing the less attentive tokens has been discussed for the image classification task but can not be extended to semantic segmentation since a dense prediction is required for every patch. To this end, this work introduces a Dynamic Token Pruning (DToP) method based on the early exit of tokens for semantic segmentation. Motivated by the coarse-to-fine segmentation process by humans, we naturally split the widely adopted auxiliary-loss-based network architecture into several stages, where each auxiliary block grades every token's difficulty level. We can finalize the prediction of easy tokens in advance without completing the entire forward pass. Moreover, we keep $k$ highest confidence tokens for each semantic category to uphold the representative context information. Thus, computational complexity will change with the difficulty of the input, akin to the way humans do segmentation. Experiments suggest that the proposed DToP architecture reduces on average $20\%\sim35\%$ of computational cost for current semantic segmentation methods based on plain vision transformers without accuracy degradation.
\end{abstract}

\section{Introduction}
\label{sec:intro}

The Transformer~\cite{vaswani2017attention} is a remarkable invention because of its exceptional capability to model long-range dependencies in natural language processing. It has been extended to computer vision applications and is known as the Vision Transformer (ViT), by treating every image patch as a token~\cite{dosovitskiy2021an}. Benefiting from the global multi-head self-attention, competitive results have been achieved on various vision tasks, \eg image classification~\cite{dosovitskiy2021an,yuan2021tokens}, object detection~\cite{carion2020end,zhu2021deformable} and semantic segmentation~\cite{cheng2022masked,cheng2021per,zhang2022segvit}. However, heavy computational overhead still impedes its broad application, especially in resource-constrained environments. In semantic segmentation, the situation deteriorates since high-resolution images generate numerous input tokens. Therefore, redesigning lightweight architectures or reducing computational costs for ViT has attracted much research attention.

\begin{figure}[t]
    \centering
    \includegraphics[width=0.98\linewidth]{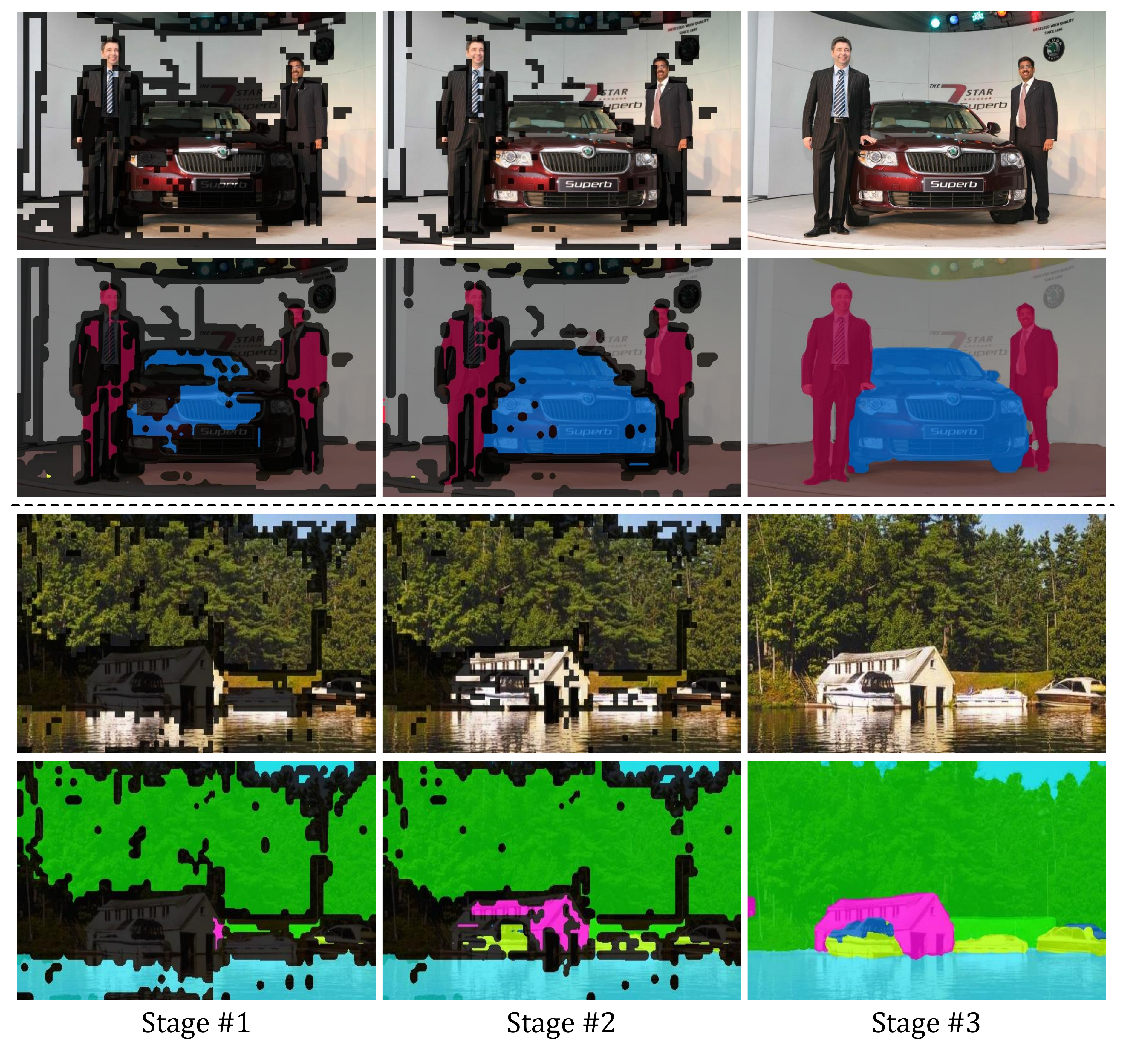}
    \caption{\textbf{Illustration of token difficulty levels by three stages using ADE20K dataset}. The network is naturally split into stages using inherent auxiliary blocks. Each sextuplet presents the early-exited/pruned tokens and their corresponding predictions successively for an image, where bright areas represent early-exited easy tokens at the current stage, while the dark ones are the kept hard tokens for the following computing.}
    \label{fig:1}
\end{figure}

Since the computational complexity of vision transformers is quadratic to the token number, decreasing its magnitude is a direct path to lessen the burden of computation. There has been a line of works studying persuasive techniques of token pruning regarding the image classification task. For example, DynamicViT~\cite{rao2021dynamicvit} determines kept tokens using predicted probability by extra subnetworks, and EViT~\cite{liang2022evit} reorganizes inattentive tokens by computing their relevance with the $[cls]$ token. Nevertheless, \textit{removing tokens, even if they are inattentive, can not be directly extended to semantic segmentation since a dense prediction is required for every image patch}. Most recently, Liang \etal~\cite{liang2022expediting} proposed a token reconstruction layer that rebuilds clustered tokens to address the issue.

In this work, a fresh angle is taken and breaks out of the cycle of token clustering or reconstruction. Motivated by humans' coarse-to-fine and easy-to-hard segmentation process, we progressive grade tokens by their difficulty levels at each stage. Hence for easy tokens, their predictions can be finalized in very early layers and their forward propagation can be halted early on. Consequently, only hard tokens are processed in the following layers. We refer to the process as the \textit{early exit} of tokens. Figure~\ref{fig:1} gives an illustration. The main body of the relatively larger objects in the image is first recognized and their process is ceased, while deeper layers progressively handle those challenging and confusing boundary regions and smaller objects. These predictions from the staged, early-exiting process can be used together with those from the completed inference. Since both outputs then form the final results jointly, it requires no token reconstruction operation and results in a simple yet effective form of efficient ViT for segmentation.

This work introduces a novel Dynamic Token Pruning (DToP) paradigm in plain vision transformers for semantic segmentation. Given that auxiliary losses~\cite{zhao2017pyramid,zheng2021rethinking} are widely adopted, DToP divides a transformer into stages using the inherent auxiliary blocks without introducing extra modules and calculations. \textit{While previous works discard auxiliary predictions irrespectively, we make good use of them to grade all tokens' difficulty levels.} The intuition of such a design lies in the dissimilar recognition difficulties of image patches represented by individual tokens. Easy tokens are halted and pruned early on in the ViT, while hard ones are kept to be computed in the following layers. We note that having this observation and shifting from auxiliary-loss-based architecture to DToP for token reduction is a non-trivial contribution. A possible situation exists where objects consisting of only extremely easy tokens, \eg sky. As a result, DToP completely discards tokens from easy-to-recognize categories in early layers, and this causes a severe loss of contextual information for the few remaining tokens in their computations. To fully utilize the inter-class feature dependencies and uphold representative context information, we keep $k$ highest confidence tokens for each semantic category during each pruning process. 

Contributions are summarized as follows:
\begin{itemize}
    \item We introduce a dynamic token pruning paradigm based on the early exit of easy-to-recognize tokens for semantic segmentation transformers. The finalized easy tokens at intermediate layers are pruned from the rest of the computation, and others are kept for continued processing.
    \item We uphold the context information by retaining $k$ highest confidence tokens for each semantic category for the following computation, which improves the segmentation performance by guaranteeing that enough contextual information is available even in extremely easy cases.
    \item We apply DToP to mainstream semantic segmentation transformers and conduct extensive experiments on three challenging benchmarks. Results suggest that DToP can reduce up to $35\%$ computation costs without a notable accuracy drop.
\end{itemize}

\section{Related Work}
\label{sec:re-work}
\subsection{Semantic Segmentation Transformers}
Semantic segmentation assigns each pixel a predefined semantic category for the purpose of pixel-level image parsing. In the last decade, deep learning techniques have considerably facilitated the development of semantic segmentation approaches. Vision transformers~\cite{dosovitskiy2021an,zhang2022segvit} now take up the baton to continue advancing the field after the great success of convolutional neural networks~\cite{he2016deep,liu2022convnet}. ViT~\cite{dosovitskiy2021an} adapts the standard Transformer~\cite{vaswani2017attention} architecture to computer vision with the fewest possible modifications, which obtains competitive results and inspires recent approaches. SETR~\cite{zheng2021rethinking} first employs ViT as an encoder and incorporates a convolutional decoder, achieving impressive performance on semantic segmentation benchmarks. SegFormer~\cite{xie2021segformer} goes beyond the plain architecture and introduces pyramid features to acquire multi-scale contexts. Segmenter~\cite{strudel2021segmenter} uses learnable class tokens as well as the output of the encoder to predict segmentation masks, which is data-dependent. SegViT~\cite{zhang2022segvit} further explores the capacity of the critical self-attention mechanism and proposes a novel attention-to-mask module to dynamically generate precise segmentation masks. In semantic segmentation, high-resolution images usually imply a large number of tokens. The attendant high computational complexity in vision transformers may be blamed for their limited applications.

\subsection{Token Reduction}
Since the computational complexity of vision transformers is quadratic to the length of input sequences, decreasing the number of tokens seems straightforward to reduce computation costs. DynamicViT~\cite{rao2021dynamicvit} observes that an accurate image classification can be obtained by a subset of most informative tokens and proposes a dynamic token specification framework. EViT~\cite{liang2022evit} demonstrates that not all tokens are attentive in multi-head self-attention and reorganizes them based on the attentiveness score with the $[cls]$ token. A-ViT~\cite{yin2022vit} computes a halting score for each token using the original network parameter and reserves computing for only discriminative tokens. 

These token reduction approaches are carefully designed for image classification based on the intuition that removing uninformative tokens (\eg backgrounds) yields a minor negative impact on the final recognition. However, things changed in semantic segmentation as we are supposed to make predictions on all image patches. Liang \etal~\cite{liang2022expediting} develop token clustering/reconstruction layers to decrease the number of tokens at middle layers and increase the number before the final prediction. 
Lu \etal~\cite{lu2023content} introduced an auxiliary policynet before transformer layers to guide the token merging operation in regions with similar content.
SparseViT~\cite{chen2023sparsevit} introduced a pruning routine on Swin Transformer for dense prediciton tasks.
Differently, we perform token reduction by finalizing the prediction of easy tokens at intermediate layers and reserving computing only for hard tokens in a dynamic manner.

\subsection{Comparisons to Prior Works}
In image classification, DVT~\cite{wang2021not} determines the patch embedding granularity and generates different token numbers based on varying recognition difficulties at the image level. Easy images can be accurately predicted with a mere number of tokens, and hard ones need a finer representation. Going one step further, we base DToP on the assumption that image patches with varying contents represented by tokens are of dissimilar recognition difficulties in semantic segmentation. We can halt easy tokens and reserve only hard tokens for subsequent computing by making early predictions via auxiliary blocks at intermediate layers. As we directly combine the early predictions for easy tokens to form the final recognition results, DToP yields no information loss during token reduction and thus requires no token reconstruction operation, compared with the method proposed by Liang \etal~\cite{liang2022expediting}.

DToP is also inspired by the deep layer cascade (LC)~\cite{li2017not} but possesses the following two unique characteristics. Firstly, DToP applies to plain vision transformers and LC pyramid convolutional neural networks. The appealing architectural properties of vision transformers enable DToP to reduce computation costs without modifying the network architecture or operators, while LC requires specific region convolution. Secondly, DToP keeps $k$ highest confidence tokens for each semantic category for the subsequent computing, which prevents easy category from halting early, contributing to the effective exploitation of contextual information.

\section{Method}
\label{sec:method}
This work introduces a Dynamic Token Pruning method based on the early exit of tokens, which expedites plain vision transformers for semantic segmentation. We detail the paradigm in this section.

\begin{figure*}[t]
    \centering
    \includegraphics[width=0.92\linewidth]{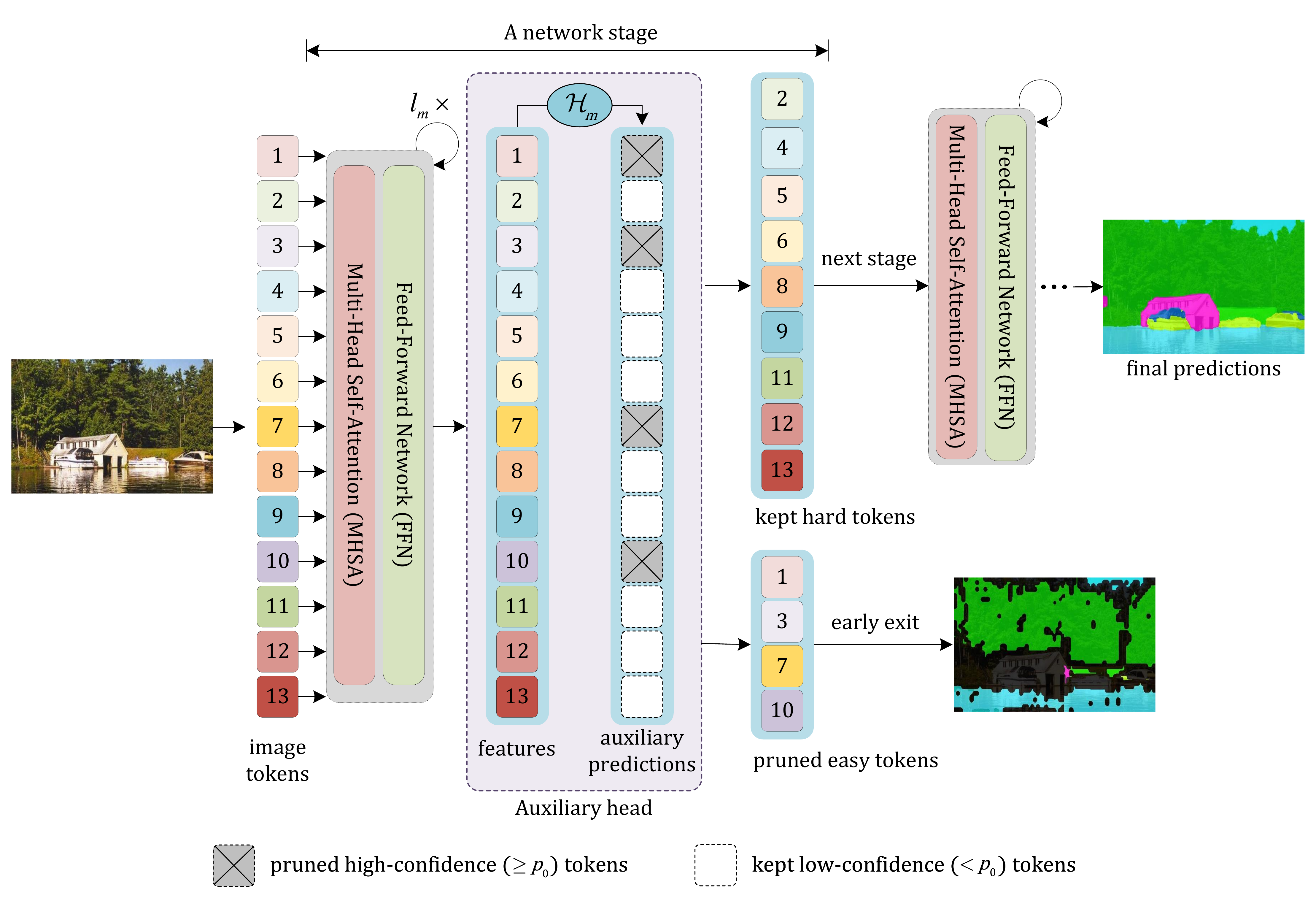}
    \caption{\textbf{Illustration of the proposed DToP framework.} Given an existing plain vision transformer, we divide it into stages using the inherent auxiliary heads. At the final layer (indexed by $l_m$) of the $m$-th stage, we use the auxiliary block $\mathcal{H}_m$ to grade all token difficulty levels. We finalize the predictions of high-confidence easy tokens at the current stage and handle other low-confidence hard tokens in the following stages. The retained $k$ highest confidence tokens for each semantic category to uphold representative context information are not presented for simplicity. Predictions from each stage jointly form the final results.}
    \label{fig:2}
\end{figure*}

\subsection{Preliminary}
A conventional vision transformer~\cite{dosovitskiy2021an} splits an image $X \in \mathbb{R}^{3 \times H \times W}$ into different patches. We then obtain a sequence of $\frac{HW}{P^2} \times C$ via patch embedding. $H$ and $W$ represent the image resolution, $P$ is the patch size and $C$ is the feature dimension. Let $N=\frac{HW}{P^2}$ be the length of the input sequence, \ie the number of tokens. Vision transformers are position-agnostic, and we generally add positional encoding to represent the spatial information of each token. The resulting sequence is denoted as $Z_0 \in \mathbb{R}^{N \times C}$, which serves as the input.

Vision transformers are usually developed from repeated units that contain a multi-head self-attention (MHSA) module and a feed-forward network (FFN). Layer normalization (LN)~\cite{ba2016layer} and residual connection~\cite{he2016deep} are employed within such units. We refer to a unit as one layer indexed by $l \in \{1,2,...,L\}$, and the output of each layer is marked as $Z_l$.
\begin{equation}
\begin{split}
    &Z_l^{\prime}=\mathtt{MHSA}(\mathtt{LN}(Z_{l-1}))+Z_{l-1},\\
    &Z_l=\mathtt{FFN}(\mathtt{LN}(Z_l^{\prime}))+Z_l^{\prime}.
\end{split}
\end{equation}
Note that FFN includes a non-linear activation function, \eg GeLU~\cite{hendrycks2016gaussian}.

\subsection{Dynamic Token Pruning}
Since a token is a natural representation of an image patch, we can finalize the prediction for easy tokens in advance without the need for complete forward computing by mimicking the segmentation process of humans. We refer to it as the \textit{early exit} of tokens, where easy tokens are halted and pruned in the early stages while hard ones are preserved for calculation at the latter stages. By doing so, fewer tokens are processed in the following layers, significantly reducing the computation costs.

As shown in Figure~\ref{fig:2}, we divide a plain vision transformer backbone into $M$ stages using its inherent auxiliary blocks $\mathcal{H}_m$ ($m \in \{1,2,...,M\}$) at the end of each stage. Let $P_m \in \mathbb{R}^{N \times K}$ represent the predicted results at the $m$-th stage, where $K$ is the number of semantic category. Suppose that tokens have finished $l_m$ layers of forward propagation at this point, then:
\begin{equation}
    P_m=\mathcal{H}_m(Z_{l_m}).
\end{equation}
$p_{m,n}$ coming from $P_m$ is the maximum predicted probability of the $n$-th token. Previous works adopt $P_m$ to calculate auxiliary losses during training and discard them irrespectively during inference. This work highlights that easy tokens can be correctly classified with high predictive confidence in these auxiliary outputs (\ie $P_m$). The proposed DToP expects to fully explore their potential ability to tell apart easy and hard tokens during both training and inference.

Inspired by~\cite{hendrycks17baseline}, we grade all token difficulty levels using $P_m$ based on a simple criterion. Assume a large confidence threshold $p_0$, \eg $0.9$. Easy tokens are classified with higher than $90\%$ scores, while hard ones are classified with lower scores. Since confident predictions for easy tokens are obtained, we prune them and halt their continued forward propagation. Hard tokens are reserved in computing in the following layers to achieve a reliable prediction. In other words, we prune the $n$-th token in $Z_{l_m}$ if $p_{m,n} \geqslant p_0$, otherwise we keep it. After propagating an image through the whole network, we combine the predicted token labels from each stage to form the final results.

\subsection{Query Matching Auxiliary Block}
Within the DToP framework, the auxiliary block for grading all token difficulty levels should follow two principles: capable of accurately estimating token difficulty levels and with a lightweight architecture. Therefore, we take the most recent attention-to-mask module (ATM)~\cite{zhang2022segvit} to achieve this goal. Specifically, a series of learnable class tokens exchange information with the encoder features using a transformer decoder. The output class tokens are used to get class probability predictions. The attention score regarding each class token is used to form a mask group. The dot product between the class probability and group masks produces the final prediction.

Two modifications are made to adapt ATM into the DToP framework. First, we decrease the number of layers in ATM as we observe no significant performance perturbation in the DToP framework with the original setting, which also guarantees a low computational overhead. Second, we decouple multiple cascaded ATM modules and use them as separate auxiliary segmentation heads, each with individual learnable class tokens. We note that we take the powerful ATM module to grade all token difficulty levels as an example, as a reliable estimation of tokens' segmentation difficulty may lead to a good accuracy-computation trade-off. Any other existing segmentation heads are of the same effect (see~\cite{xie2021segformer,zhao2017pyramid,zheng2021rethinking} for examples). In Section~\ref{sec:exp}, we also provide experiments with the regular FCN head~\cite{long2015fully} to validate the generality of DToP.

\subsection{Upholding Context Information}
Scenarios exist where all tokens of a specific semantic category are extremely easy to recognize, \eg sky. Such tokens may be entirely removed in early layers, resulting in a loss of context information in the following layers of calculation. Practices~\cite{yu2020context,yu2018learning} indicate that fully exploiting the inter-category contextual information improves the overall semantic segmentation accuracy. To this end, we keep $k$ highest confidence tokens for each semantic category during each pruning process. \textit{Only the categories that appear in the current image are considered}. Given a specific semantic category, if the number of tokens with a higher than $p_0$ score is more than $k$, then the top-$k$ of them are kept. Otherwise, we keep the actual number of them. These category-known tokens join in the calculation along with other low-confidence ones, so semantic information of easy category is preserved for inter-category information exchange, leading to an accurate semantic segmentation.

\section{Experiments}
\label{sec:exp}

\subsection{Datasets and Metrics}
\textbf{ADE20K}~\cite{zhou2017scene} is a widely adopted benchmark dataset for semantic segmentation. It contains about $20k$ images for training and $2k$ images for validation. All images are labeled with $150$ semantic categories. \textbf{COCO-Stuff-10K}~\cite{caesar2018coco} dataset contains $9k$ images for training and $1k$ images for testing. Following~\cite{zhang2022segvit}, we use $171$ semantic categories for experiments. \textbf{Pascal Context}~\cite{mottaghi2014role} has a total of $10,100$ images, of which $4,996$ images are for training and $5,104$ for validation. It provides pixel-wise labeling for $59$ categories, excluding the background.

Following the common convention, we use the mean intersection over union (mIoU) to evaluate the segmentation accuracy and the number of float-point operations (FLOPs) to estimate the model complexity. The computation in DToP is unevenly allocated among easy and hard samples by pruning different numbers of tokens. We thus report the average FLOPs over the entire validation/test dataset.

\subsection{Implementation Details}
We adopt the plain vision transformer incorporating the adapted ATM module as the baseline model, where ATM modules work as auxiliary heads. We follow the standard training settings in \textit{mmsegmentation}\footnote{\url{https://github.com/open-mmlab/mmsegmentation}} and use the same hyperparameters as the original paper. All reported mIoU scores are based on single-scale inputs. $k$ is set to $5$ in this work. As changing $p_0$ within a certain range ($0.90\sim0.98$) during training leads to similar results, we empirically fix it to $0.95$ for all training processes unless specified.

\subsection{Ablation Study}
We first conduct extensive ablation studies with the ADE20K dataset~\cite{zhou2017scene} using ViT-Base~\cite{dosovitskiy2021an} as the backbone.

\subsubsection{Necessity for Model Training}
Using auxiliary heads for efficient training is a common convention in the semantic segmentation community, see~\cite{cheng2022masked,zhao2017pyramid,zheng2021rethinking} for examples. Generally, the auxiliary outputs are discarded at test time. As the proposed DToP grades all token difficulty levels using the auxiliary outputs, we can apply DToP to existing methods off-the-shelf during inference. Therefore, we verify the necessity for model retraining or finetuning under the proposed DToP framework. We denote DToP@Direct as directly applying DToP to the baseline model during inference. DToP@Finetune means finetuning the segmentation heads for $40k$ iterations on the baseline model using DToP, and DToP@Retrain retraining the entire model using DToP for $160k$ iterations.

Results are shown in Table~\ref{tab:1}. We observe that all three settings reduce the computation costs by about $20\%$, where DToP@Direct and DToP@Retrain lead to a significant accuracy drop while DToP@Finetune performs slightly better. Results suggest that the proposed DToP@Finetune requires only a little extra training time but significantly reduces the computational complexity while maintaining accuracy. We adopt the @Finetune setting in the following experiments. Note that the slight fluctuation in FLOPs of the three training schemes comes from varied predictions of auxiliary heads in the individual training processes.

\begin{table}
    \begin{center}
    \begin{tabular}{lcl}
    \toprule
    \textbf{Method} & \textbf{GFLOPs} & \textbf{mIoU(\%)} \\
    \midrule\midrule
    Baseline & 109.9 & 49.7 \\
    + DToP@Direct ($\clubsuit .0$) & 87.5 & 47.9 (-1.8) \\
    + DToP@Finetune ($\clubsuit 2.5$) & 86.8 & \textbf{49.8} (+0.1) \\
    + DToP@Retrain ($\clubsuit 12.0$) & 87.5 & 49.1 (-0.6) \\
    \bottomrule
    \end{tabular}
    \end{center}
    \caption{\textbf{Comparison of training schemes.} With a short finetuning scheme, the pruned model achieves even better results than the baseline.  $\clubsuit$ means extra training time in hours on 8 NVIDIA A100 cards.}
    \label{tab:1}
\end{table}

\subsubsection{Ablation for Confidence Threshold}
The confidence threshold $p_0$ is a crucial hyperparameter that decides the pruned token number in each pruning process and directly affects the trade-off between computation cost and accuracy. Quantitative results are shown in Table~\ref{tab:2}. When $p_0=1$, the model degenerates to the baseline architecture. As $p_0$ decreases, more easy tokens are pruned as well as more unreliable early predictions. We observe that the performance saturates at $p_0=0.95$ when using ATM as the segmentation head. 

\begin{figure*}[t]
    \centering
    \includegraphics[width=.98\linewidth]{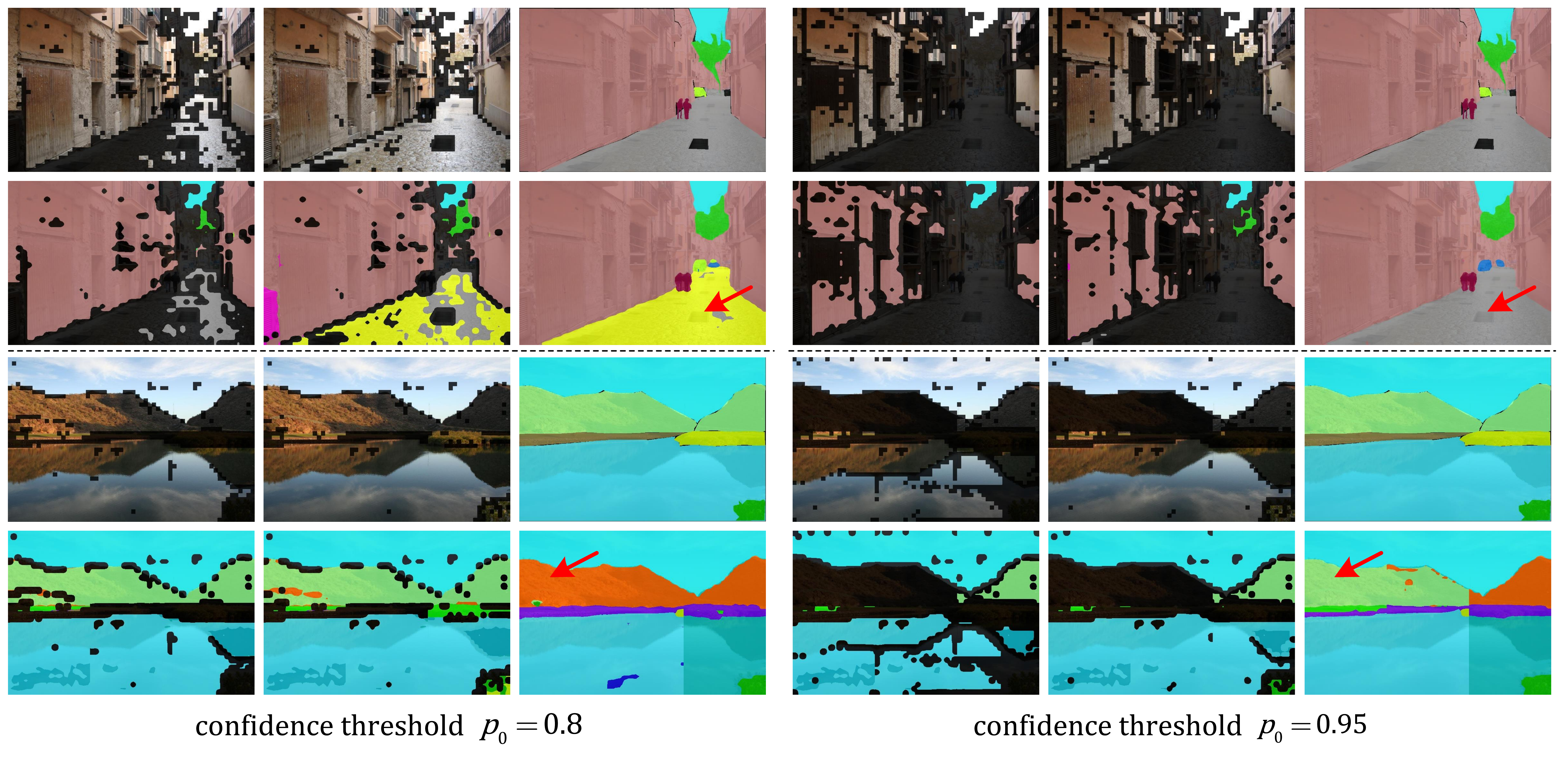}
    \caption{\textbf{Illustration of the effects for different confidence threshold}. Samples are from ADE20K dataset. For each sextuplet, we show the pruned token distribution and the ground truth (first row), as well as its corresponding segmentation results (second row). Bright areas represent pruned tokens, and those in the dark are kept tokens for the following computing. A small $p_0$ value (left two examples) leads to more pruned tokens in early stages but results in inferior segmentation results (see the red arrow). }
    \label{fig:3}
\end{figure*}

We also verify the value using SETR~\cite{zheng2021rethinking} (w/ the naive segmentation head described in FCN~\cite{long2015fully}) and show the results in Table~\ref{tab:3}. We observe that for FCN head $p_0=0.98$ may be a better choice. In practice, the value can be chosen empirically with a small validation set. We also note that for SETR, DToP@Direct has already obtained a promising mIoU score of $46.6\%$ that is only $0.4\%$ lower than the baseline but with significantly reduced computation ($\sim 23.4\%$). Some qualitative examples of how the threshold $p_0$ affects the pruned token number and segmentation accuracy are shown in Figure~\ref{fig:3}\footnote{Note that some pruned tokens change their final segmentation due to the attention-to-mask mechanism in ATM but will remain the same in regular FCN heads.}.

\begin{table}
    \begin{center}
    \resizebox{1.\linewidth}{!}{
    \begin{tabular}{c|ccccccc}
       \toprule
       \textbf{$\boldsymbol{p_0}$} & 0.60 & 0.70 & 0.80 &0.85 & 0.90 & 0.95 & 1.00 \\
       \midrule
       \textbf{GFLOPs} & 70.2 & 73.4 & 77.8 & 80.7 & 83.6 & 86.8 & 109.9 \\
       \midrule
       \textbf{mIoU(\%)}& 46.8 & 48.0 & 49.0 & 49.3 & 49.5 & \textbf{49.8} & 49.7 \\
       \bottomrule
    \end{tabular}}
    \end{center}
    \caption{\textbf{Ablation for confidence threshold $p_0$}.
    The results are evaluated on ADE20K with ATM head.}
    \label{tab:2}
\end{table}

\begin{table}
    \begin{center}
    \begin{tabular}{lccl}
       \toprule
       \textbf{Method} & $\boldsymbol{p_0}$ & \textbf{GFLOPs} & \textbf{mIoU(\%)} \\
       \midrule\midrule
       SETR & - & 107.7 & 47.0 \\
       + DToP@Direct & 0.90 & 74.0 & 45.6 (-1.4)\\
       + DToP@Finetune & 0.90 & 72.5 & 46.3 (-0.7)\\
       \midrule
       + DToP@Direct & 0.95 & 78.3 & 46.2 (-0.8) \\
       + DToP@Finetune & 0.95 & 76.5 & 46.8 (-0.2) \\
       \midrule
       + DToP@Direct & 0.98 & 82.5 & \underline{46.6} (-0.4) \\
       + DToP@Finetune & 0.98 & 80.6 & \textbf{47.0} (+0.0) \\
       \bottomrule
    \end{tabular}
    \end{center}
    \caption{\textbf{Ablation results based on SETR.} About $25\%$ of the tokens can be pruned with no performance dropped.}
    \label{tab:3}
\end{table}

\begin{figure*}[t]
    \centering
    \includegraphics[width=.98\linewidth]{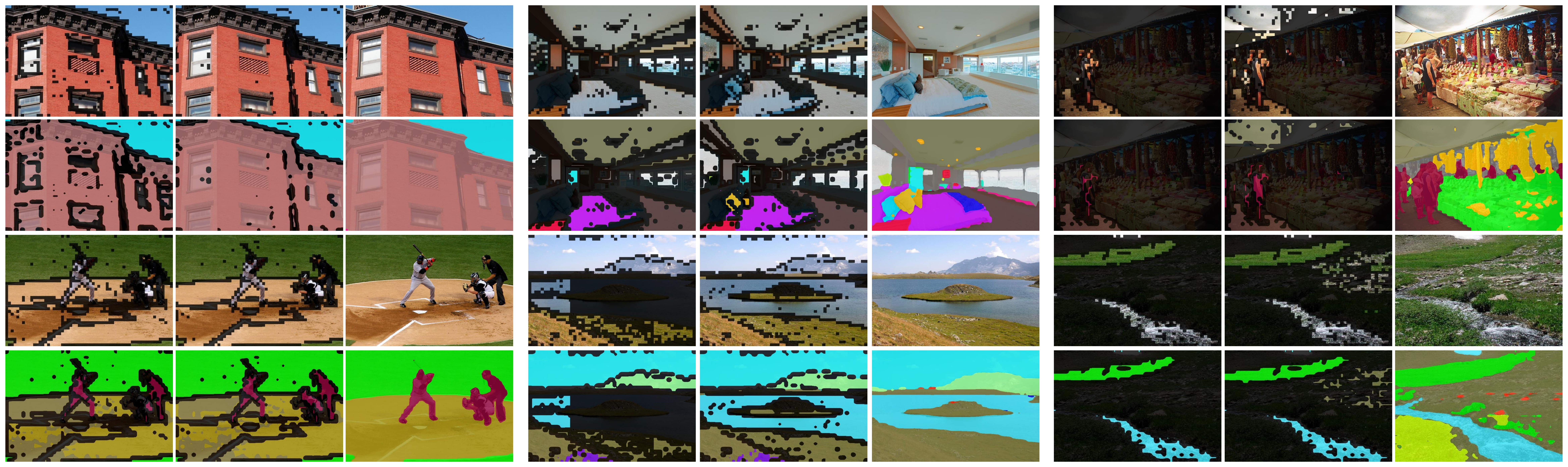}
    \caption{\textbf{Prediction results of three stages during the token pruning processes.} Examples from ADE20K with different image complicity: most tokens are pruned (left group), the majority are pruned (middle group), and very few are pruned (right group). For each sextuplet, we show the pruned token distribution and the corresponding segmentation results at each stage.}
    \label{fig:4}
\end{figure*}
\vspace{-0.5em}
\subsubsection{Exploration on Pruning Position}
The critical insight of DToP is to finalize the prediction of easy tokens in intermediate layers and prune them in the following calculation by grading all tokens' difficulty levels. Thus the position of auxiliary heads matters. It affects the recognition accuracy of pruned easy tokens and the trade-off between computation cost and segmentation accuracy. We conduct explorations on the pruning position $l_m$ and show the results in Table~\ref{tab:4}. Results demonstrate that dividing the backbone into three stages with token pruning at the $6^{th}$ and $8^{th}$ layers achieves an expected trade-off between computation cost and segmentation accuracy. We adopt this setting in all other experiments and note that it may not be optimal on account of limited explorations.

\begin{table}[]
    \begin{center}
    \begin{tabular}{cccc}
       \toprule
       \textbf{Stages} & \textbf{Position} & \textbf{GFLOPs} & \textbf{mIoU (\%)} \\
       \midrule\midrule
       1 & Baseline & 109.9 & 49.7 \\
       2 & \{6\} & 85.7 & 49.4 \\
       2 & \{8\} & 92.1 & 49.4 \\
       3 & \{6, 8\} & 86.8 & \textbf{49.8} \\
       4 & \{3, 6, 8\} & 74.5 & 48.3 \\
       \bottomrule 
    \end{tabular}
    \end{center}
    \caption{\textbf{Exploration of the pruning position.} The first column indicates the number of divided stages.}
    \label{tab:4}
\end{table}

\begin{table}[t]
     \begin{center}
     \begin{tabular}{lccc}
        \toprule
        \textbf{Method} & \textbf{Context} & \textbf{GFLOPs} & \textbf{mIoU(\%)} \\
        \midrule\midrule
        Baseline & - & 109.9 & 49.7 \\
        Remove & $\times$ & 82.6 & 48.7 \\
        Top-$35\%$ & $\times$ & 84.6 & 48.7 \\
        Average & $\checkmark$ & 83.5 & 48.9 \\
        Top-$k$ & $\checkmark$ & 83.6 & 49.5 \\
        Avg \& Top-$k$& $\checkmark$ & 83.6 & \textbf{49.7} \\
        \toprule
     \end{tabular}
     \end{center}
     \vspace{-0.5em}
     \caption{\textbf{Comparison of different pruning methods.} All models are trained with DToP@Finetune using $p_0=0.9$.}
     \label{tab:5}
\end{table}

\begin{table}[H]
    \begin{center}
    \resizebox{.95\linewidth}{!}{
    \begin{tabular}{lcccc}
       \toprule
       \textbf{Method} & \textbf{Decode} &\textbf{Aux} &  \textbf{GFLOPs} & \textbf{mIoU (\%)} \\
       \midrule\midrule
       Baseline & ATM & ATM & 109.9 & 49.7 \\
       + DToP@Finetune & ATM & ATM & 83.6 & 49.5 \\
       \midrule
       Baseline & FCN & FCN & 107.7 & 47.0 \\
       + DToP@Finetune & FCN & FCN & 80.6 & 47.0 \\
      \midrule
       Baseline & FCN & ATM & 107.7 & 49.6 \\
       + DToP@Finetune & FCN & ATM & 83.4 & 48.4 \\
       \midrule
       Baseline & ATM & FCN & 109.9 & 47.9 \\
       + DToP@Finetune & ATM & FCN & 73.3 & 46.9 \\
       \bottomrule
    \end{tabular}}
    \end{center}
    \caption{\textbf{Exploration of different segmentation heads.} Results in the second part uses $p_0=0.98$ and others $0.9$. `Decode' means the final decoder head and `Aux' auxiliary head.}
    \label{tab:6}
\end{table}

\subsubsection{Ablation for Pruning Method}
After grading all token difficulty levels at the current stage, the specific pruning method is flexible. We experiment with four token pruning methods. Following LC~\cite{li2017not}, we remove easy tokens directly without the consideration of halting easy category information. In contrast, this work keeps $k$ highest confidence tokens for each appeared semantic category to uphold representative context information, marked as top-$k$. An alternative to uphold context information is to average all easy token values into one token for each semantic category. We also prune a fixed proportion of tokens by removing the top $35\%$ highest confidence tokens to evenly allocate computation among images. Results are shown in Table~\ref{tab:5}, where the proposed top-$k$ method outperforms others by a large margin, suggesting its effectiveness. Furthermore, we observe that methods upholding context information, \ie the Average and top-$k$, are superior to others.
\vspace{-0.5em}
\subsubsection{Influence of Segmentation Heads}
In Table~\ref{tab:6}, we verify different segmentation heads and observe that the proposed DToP performs effectively in both ATM and FCN settings (first two parts), indicating its general applicability.
To ensure a fair comparison, we selected different $p_0$ values to maintain similar GFlops and compared the performance. We noticed that the choice of the auxiliary head significantly influences the performance. Particularly, the powerful ATM head provides a more accurate estimation of all tokens' difficulty levels, resulting in superior results.

\begin{table*}[t]
    \begin{center}
    \small
    \resizebox{.95\linewidth}{!}{
    \begin{tabular}{lcc|cc|cc|cc}
       \toprule
       \multirow{2}{*}{\textbf{Method}} & \multirow{2}{*}{\textbf{Backbone}} & \multirow{2}{*}{$\boldsymbol{p_0}$} & \multicolumn{2}{c|}{\textbf{ADE20K}} & \multicolumn{2}{c|}{\textbf{Pascal Context}} & \multicolumn{2}{c}{\textbf{COCO-Stuff-10K}} \\
       ~ & ~  & ~ & \textbf{mIoU(\%)} & \textbf{GFLOPs} & \textbf{mIoU(\%)} & \textbf{GFLOPs} & \textbf{mIoU(\%)} & \textbf{GFLOPs} \\
       \midrule\midrule
       SETR~\cite{zheng2021rethinking} & ViT-Base & - & 47.0 & 107.7 & 58.1 & 92.4 & 41.2 & 107.7 \\
       + DToP@Finetune & ViT-Base & 0.90 & 46.3 & 72.5 & 57.5 & 61.4 & 40.6 & 77.6 \\
       + DToP@Finetune & ViT-Base & 0.98 & \underline{47.0} & 80.6 & \textbf{58.2} & 69.1 & \underline{40.9} & 86.4 \\
       
       \midrule
       SegViT~\cite{zhang2022segvit} & ViT-Large & - & 53.3 & 617.0 & 63.0 & 315.4 & 47.4 & 366.9 \\
       + DToP@Finetune & ViT-Large & 0.90 & 52.4 & 380.3 & 62.2 & 206.1 & 46.6 & 253.1 \\
       + DToP@Finetune & ViT-Large & 0.95 & \underline{52.8} & 412.8 & \underline{62.7} & 224.3 & \underline{47.1} & 276.2 \\
       \bottomrule
    \end{tabular}}
    \end{center}
    \caption{\textbf{Main results on three semantic segmentation benchmarks.} We apply the proposed DToP with the finetuning training scheme to current state-of-the-art semantic segmentation networks based on plain vision transformers. GFLOPs is the average number of the whole validation dataset. We perform token pruning at $\{8^{th}, 16^{th}\}$ layers for ViT-Large.}
    \label{tab:7}
\end{table*}

\begin{figure*}[t]
    \centering
    \includegraphics[width=0.9\linewidth]{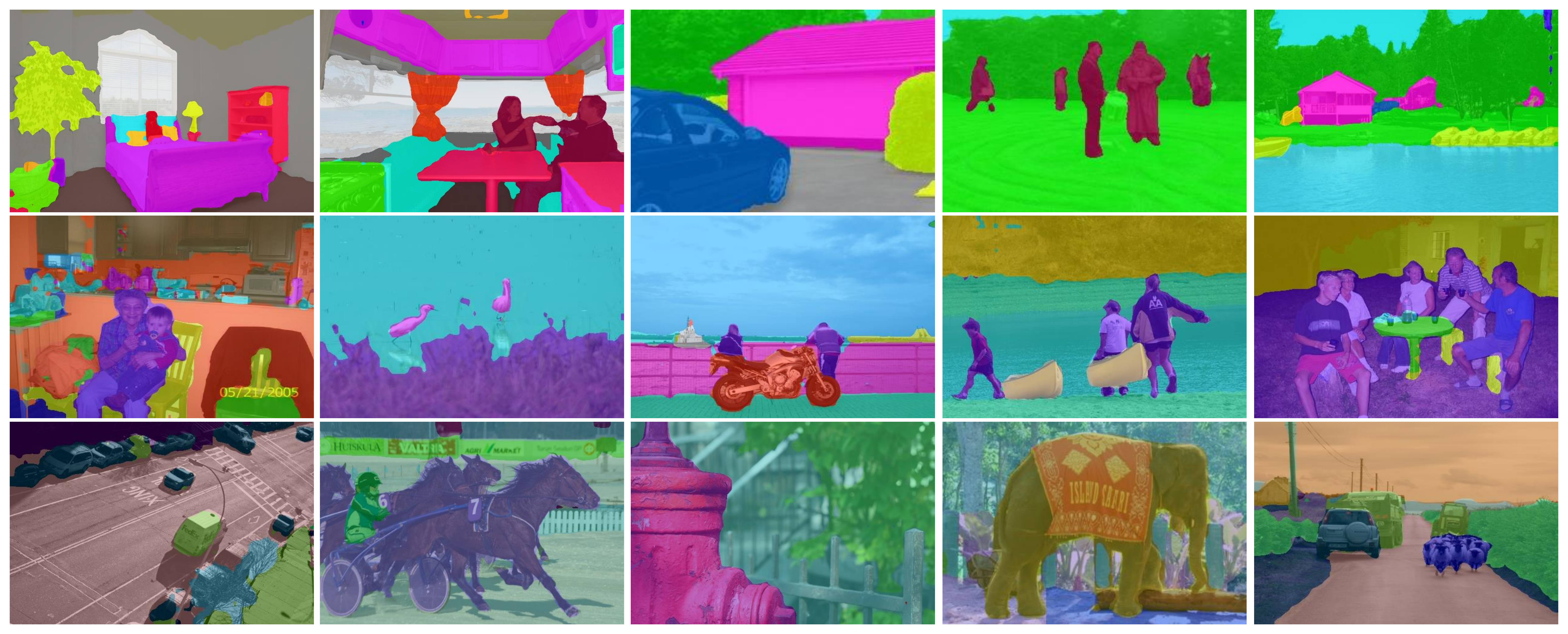}
    \caption{\textbf{Visualized results.} The segmentation results are predicted on ADE20K (first row), Pascal Context (middle row), and COCO-Stuff-10K (last row). The model is SegViT with DToP@Finetune based on ViT-Large.}
    \label{fig:5}
\end{figure*}

\subsection{Application to Existing Methods}
We apply the proposed DToP to two mainstream semantic segmentation frameworks in plain vision transformers~\cite{dosovitskiy2021an}. SETR~\cite{zheng2021rethinking} uses the naive upsampling decoder, and SegViT~\cite{zhang2022segvit} adopts our adapted ATM module. Results are shown in Table~\ref{tab:7} using three challenging benchmarks. With an appropriate confidence threshold $p_0$, the proposed DToP can reduce on average $20\%\sim35\%$ computation cost without notable accuracy degradation. More specifically, SETR with DToP@Finetune reduces $25.2\%$ computation cost (FLOPs $107.7$G $\rightarrow$ $80.6$G) without mIoU drop on ADE20K and even obtains a slightly better mIoU ($58.1\% \rightarrow 58.2\%$) on Pascal Context dataset. SegViT with DToP@Finetune based on ViT-large reduced about $35\%$ computation with only $0.5\%$ mIoU lower on ADE20K.

A qualitative comparison regarding the pruned token number of different images is presented in Figure~\ref{fig:4}. We see that most tokens are pruned at very early stages for images of simple scenarios. For complex scene images, most tokens remain until the final prediction. Consequently, the computation is unevenly allocated among images by adjusting the pruned token number, yielding a considerable improvement in computation efficiency. We also observe that pruned easy tokens are primarily located at the central area of objects, while kept hard tokens are located on the boundaries, similar to the segmentation process by humans. Some visualized predictions are shown in Figure~\ref{fig:5}.

\vspace{-0.3em}
\section{Conclusion}
\label{sec:conc}
This work studies the problem of reducing computation costs for existing semantic segmentation based on plain vision transformers. A Dynamic Token Pruning paradigm is proposed based on the early exit of tokens. Motivated by the coarse-to-fine segmentation process by humans, we assume that different tokens representing image regions have dissimilar recognition difficulties and grade all tokens' difficulty levels using the inherent auxiliary blocks. To this end, we finalize the predictions of easy tokens at intermediate layers and halt their forward propagation, which dynamically reduces computation. We further propose a strategy to uphold context information by preserving extremely easy semantic categories after token pruning. Extensive experimental results suggest that the proposed method achieves compelling performance.

Similar to all other dynamic networks, DToP can not take full advantage of the calculation efficiency of a mini-batch. We will make optimization in the future and further expedite vision transformers using the proposed DToP.

{\small
\bibliographystyle{ieee_fullname}
\bibliography{egbib}
}

\end{document}